# Efficient Learning of PDEs via Taylor Expansion and Sparse Decomposition into Value and Fourier Domains

Md Nasim, Yexiang Xue
Department of Computer Science, Purdue University, West Lafayette, IN, USA
mnasim@purdue.edu, yexiang@purdue.edu

## Abstract

Accelerating the learning of Partial Differential Equations (PDEs) from experimental data will speed up the pace of scientific discovery. Previous randomized algorithms exploit sparsity in PDE updates for acceleration. However such methods are applicable to a limited class of decomposable PDEs, which have sparse features in the value domain. We propose REEL, which accelerates the learning of PDEs via random projection and has much broader applicability. REEL exploits the sparsity by decomposing dense updates into sparse ones in both the value and frequency domains. This decomposition enables efficient learning when the source of the updates consists of gradually changing terms across large areas (sparse in the frequency domain) in addition to a few rapid updates concentrated in a small set of "interfacial" regions (sparse in the value domain). Random projection is then applied to compress the sparse signals for learning. To expand the model applicability, Taylor series expansion is used in REEL to approximate the nonlinear PDE updates with polynomials in the decomposable form. Theoretically, we derive a constant factor approximation between the projected loss function and the original one with poly-logarithmic number of projected dimensions. Experimentally, we provide empirical evidence that our proposed REEL can lead to faster learning of PDE models (70%-98% reduction in training time when the data is compressed to 1% of its original size) with comparable quality as the non-compressed models.

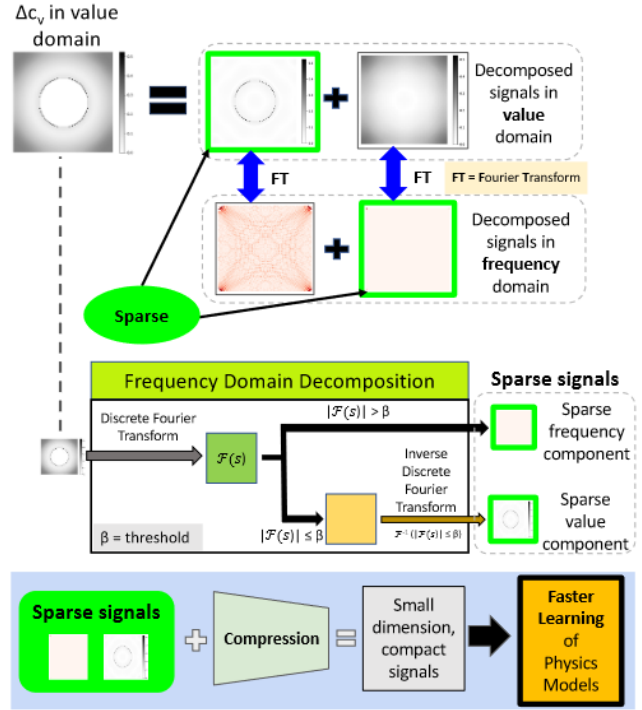

Figure 1: High-level idea of our REEL algorithm. REEL decomposes spatial and temporal updates into sparse signals in the value and the frequency domains. Random projection compresses these sparse signals for accelerated learning of PDEs. This example uses the decomposition of the vacancy concentration $c_v$ in learning the nanovoid evolution in materials under extreme conditions.

## 1 Introduction

Physics models encoded in Partial Differential Equations (PDE) are crucial in the understanding of many natural phenomena such as sound, heat, elasticity, fluid dynamics, quantum mechanics etc. Learning these physics models from data is essential to enhance our understanding of the world, accelerate scientific discovery, and design new technologies. Deep neural net-works (e.g., physics-informed neural nets Raissi et al. [2019a], Hamiltonian neural nets Greydanus et al. [2019]) have been successfully deployed in this domain. In these approaches, the spatial and temporal updates of a PDE model are matched with ground truth experimental observations. A loss function is defined based on the mismatch between the simulation and the ground truth, and then the physics model is learned by back-propagation of error gradients of the loss function Xue et al. [2021a], Raissi et al. [2019a]. Nevertheless, such learning processes are expensive because of the need to back-propagate gradients over spatial and temporal simulations involving millions of mutually interacting elements.

One line of successful approaches to accelerate the learning of PDE models exploits the sparsity nature of system changes over time. For example, during the microstructure evolution of many engineering materials, only the boundary of the microstructure changes while large portion of the system remains unchanged. It is also assumed that the corresponding PDE models can be decomposed into affine function of parameter functions and feature functions Nasim et al. [2022], Sima and Xue [2021]. The combination of decomposablity of the PDE model and sparse changes/updates over time together create opportunities for efficient algorithms which handle learning in compressed spaces using random projections and/or locality sensitive hashing. Nevertheless, such decomposablity structure applies to a limited class of PDEs and sparsity structures may change with varying initial and boundary conditions (BC/IC).

This paper propose a more general approach for efficiently learning PDE models via random projection, by exploiting sparsity in both value domain and frequency domain, and also approximating non-decomposable functions with decomposable polynomials. We observe that, systems modeled by PDEs often have slow and gradual updates across wide regions in addition to a few rapid changes concentrated in small "interfacial" regions. Such systems are frequently found in the real world. For example, during manufacturing processes such as laser sintering of powder materials into dense solids, grain boundary changes sharply at the interface area (sparse local change), while temperature rises gradually around the whole material (dense global change).

Systems with dense global change and sharp interface change limit the application of existing approaches Nasim et al. [2022], Sima and Xue [2021] for efficiently learning relevant PDE models. However, we observe that these temporal change signals *can again become sparse if they are decomposed in value and frequency domains*. The Fourier uncertainty principle Folland and Sitaram [1997] implies that a signal sparse in the value domain should be dense in the frequency domain and vice versa. As a result, this decomposition *can capture the sparse side of signals, whether they are in the value or the frequency domain.*

We propose R̲andom Projection based E̲fficient L̲earning (REEL), a general approach to expedite the learning of PDEs, using signal decomposition into value and frequency domains, polynomial approximation with Taylor series, and compression via random projection. The key innovation of REEL is the inclusion of a signal decomposition step in the PDE learning framework. With this step, we convert dense value domain system updates into sparse signal components in the value and frequency domains. We also use polynomial approximation with Taylor series to approximate PDE models, which otherwise cannot be written in the decomposable form of parameter functions and feature functions. An example is the phase field model of sintering of powder compacts Zhang and Liao [2018]. After decomposition, the sparse signal components in the value and frequency domains are compressed to smaller dimensions by random projection. The learning of PDE models is then carried out in the compressed space. Notice that both the signal decomposition and compression steps are carried out once as a preprocessing step and in parallel, thus adding little computation overhead. An overview of REEL is shown in Figure 1.

Theoretically, we show the sparse projection into the value and frequency domains biases learning in a limited way. We derive a constant factor approximation bound between the projected loss function and the original one with poly-logarithmic number of projected dimensions. Experimentally, we evaluate our approaches in several real-world problems. The first is laser sintering of materials, which involves both grain boundary and temperature changes. A second application is nanovoid defect evolution in materials under irradiation and high temperature, in which both void surface movement and the emergence of interstitial and vacancy densities are considered. We demonstrate that using our REEL algorithm leads to $70-98\%$ reduction in training times when the data is compressed to 1% of its original size and the learned models' performance are comparable to baseline.

Our contributions can be summarized as follows : 1) we propose an efficient method to learn PDEs that have both sparse and dense feature functions, 2) we extend the applicability of random projection on sparse functions to both sparse and dense functions, using an appropriate decomposition of representation into both the value and frequency domains, 3) we extend the applicability of random projection for learning PDE models that are not readily decomposable, by using Taylor series approximation, and 4) we show empirical evidence that our learning method REEL can greatly accelerate the current state of PDE model learning.

## 2 Background

**Partial Differential Equation (PDE) models.** PDE models of different orders appear in various scientific fields. Here, order refers to the highest derivative that appears in the PDE. Although our method is in principle applicable to PDEs of any orders, for simplicity we keep our discussion limited to the following formulation of PDEs with first order time derivative:

$$\frac{\partial u(\vec{p}, t)}{\partial t} = F(u, \nabla u, \nabla^2 u, \ldots, \theta) \quad (1)$$

Here, $u(\vec{p}, t)$ is a function of space $\vec{p}$ and time $t$, $\theta$ is a set of scalars and $F$ is a function of both $u$ and $\theta$, and contains spatial derivative terms such as $\nabla, \nabla^2, \ldots$ representing first, second and even higher order spatial derivatives. $u$ can be system state variable such as concentration, temperature etc., $\theta$ can be some system specific properties such as gradient coefficients, mobility parameter for a particular material etc. Different forms of $F$ denotes different system models, and for a particular model, different values of $\theta$ leads to different system dynamics.

**Learning PDE models from data.** Suppose we have a sequence of ground truth PDE trajectories $u^{GT}(t)$ for time $t = 1, 2, \ldots, T$, extracted from data, and a general form of PDE model as given in Equation 1, parameterized by $\theta$. Our goal is to learn these $\theta$ parameters.

PDE model parameters $\theta$ can be learned via numerical simulation Xue et al. [2021a]. To do so, we first replace the PDE derivatives with finite difference quotients i.e. $\frac{\partial u}{\partial t} \approx \frac{\Delta u}{dt}$ in the original equation. At any time $t$, we can compute the model predicted system change $\Delta u(t)$ within time interval $dt$, by solving Equation 1 parameterized by $\theta$. Additionally, From data, we can extract the ground truth system state change $\Delta u^{GT}(t) = u(t + dt) - u(t)$. Thus, we have the following at hand:

- $\Delta u^{GT}(t)$, ground truth system state change,

- $\Delta u(t)$, predicted system change from PDE model

With ground truth values of $\theta$, both $\Delta u^{GT}(t)$ and $\Delta u(t)$ should match together. Hence we can define a loss function as follows:

$$\min_\theta \ L(\theta) = \sum_{t=1}^{T} ||\Delta u(t) - \Delta u^{GT}(t)||_2^2. \quad (2)$$

The loss function in Equation 2 penalizes the difference between simulation output and ground truth observation. With this neural PDE model, we can now backpropagate the error gradients from Equation 2 and update $\theta$ using stochastic gradient descent (SGD). When learning converges and the loss becomes sufficiently small, the model discovers a set of parameters $\theta$ which leads to similar dynamics as the empirical observations.

**Efficient learning of Sparse and Decomposable PDE.** Sparse and Decomposable PDEs (SD-PDE) are a special class of PDEs, where temporal updates of the PDE model (i.e. $\frac{\partial u}{\partial t}$ in Equation 1) can be formulated as combination of sparse feature functions of system state variables $u$, multiplied by functions of learnable PDE model parameters $\theta$. This special class of PDEs is first introduced in Nasim et al. [2022], and the authors proposed RAPID-PDE algorithm, to accelerate the learning of corresponding PDE model parameters $\theta$ using random projection.

To learn SD-PDE models efficiently using random projection, a one-time projection of feature functions and system changes is performed. This random projection compresses the high-dimensional signals (features and system changes) into compressed low-dimensional space. Training epochs are then carried out in this compressed space. The sparsity of the high dimensional signals makes it possible to represent them with compact low dimensional signals, and the reduction of data dimension greatly accelerates the discovery of PDE model parameters.

**Limitations of RAPID-PDE.** Although RAPID-PDE can be highly efficient in learning PDE models,

its applicability is limited by the decomposability and sparsity structure of the PDE models. Feature functions of the PDE model instances can contain dense value domain signals, depending upon initial and boundary conditions. Moreover, not all PDE models can be decomposed readily into inner products of feature functions and parameter functions.

## 3 REEL: Efficient learning of PDEs with polynomial approximation and signal decomposition into value and frequency domains

Our REEL algorithm aim to accelerate learning of a broad class of PDE models, which 1) may not be readily decomposable into inner product parameter function and feature functions, and/or 2) may have feature functions/system changes that are not sparse in value domain. Before moving to the details of our REEL algorithm, let us first take a close look at the formulation of SD-PDE models we mention in Section 2, which have the following general format:

$$\frac{\partial u(\vec{p}, t)}{\partial t} = \vec{\phi}(\theta)\vec{W}(u) = [\phi_1(\theta), \phi_2(\theta), \ldots, \phi_n(\theta)] \begin{bmatrix} W_1(u) \\ W_2(u) \\ \vdots \\ W_n(u) \end{bmatrix} \quad (3)$$

Here, $u$ is the system state variable of interest, $W_i$ are feature functions of $u$, often sparse in value domain, and independent of PDE model parameters $\theta$. Similarly, $\phi_i$ are functions of $\theta$ and independent of $u$. Exact forms of $\phi_i$ and $W_i$ depend on the PDE model. By replacing derivatives with finite difference quotients, and using a random matrix $\boldsymbol{P}$ for random projection on both sides, we can rewrite Equation 3 as follows:

$$\frac{\boldsymbol{P}\Delta u}{dt} = \boldsymbol{P}[\phi_1(\theta), \phi_2(\theta), \ldots, \phi_n(\theta)] \begin{bmatrix} W_1(u) \\ W_2(u) \\ \vdots \\ W_n(u) \end{bmatrix}$$

$$\Rightarrow \boldsymbol{P}\Delta u = dt.[\phi_1(\theta), \phi_2(\theta), \ldots, \phi_n(\theta)] \begin{bmatrix} \boldsymbol{P}W_1(u) \\ \boldsymbol{P}W_2(u) \\ \vdots \\ \boldsymbol{P}W_n(u) \end{bmatrix} \quad (4)$$

where, $\boldsymbol{P}\Delta u$ is the *predicted* compressed system state changes, and $[\boldsymbol{P}W_1(u), \ldots, \boldsymbol{P}W_n(u)]$ are the compressed features. The ground truth system state change $\Delta u^{GT}$ is also compressed to $\boldsymbol{P}\Delta u^{GT}$ by one-time random projection with $\boldsymbol{P}$. RAPID-PDE algorithm then learns $\theta$ in the compressed space – minimizing a loss function that penalizes the difference between $\boldsymbol{P}\Delta u$ and $\boldsymbol{P}\Delta u^{GT}$.

Our REEL learning algorithm is inspired from a few observations. **First**, we notice that although it is not possible to decompose all PDE models as in Equation 3, non-linear functions can be approximated by polynomials with Taylor series expansion and then written in decomposable form. For example, $\sin(u\theta) \approx u\theta - \frac{(u\theta)^3}{3!} + \frac{(u\theta)^5}{5!} - \frac{(u\theta)^7}{7!}$, and this approximation can be written similar to Equation 3. **Second**, for decomposed PDE models, the system change $\Delta u$ and the feature functions $W_i$ in Equation 3 may not be sparse in value domains; however, a change of representation domains, i.e. combination of value domain and frequency domain can make these signals sparse. For example, in the application domain considered in this work, the sintering of powder particles with high energy heat source such as laser, heat gets diffused into the particles and surroundings, and the particles fuse together. During this sintering process, changes in the particles happen at the boundary (sparse value domain update), while the change in specimen temperature due to heat diffusion is more widespread across the whole specimen (dense value domain update). In practice, the dense temperature change, and similarly many other dense value domain signals can be represented by sparse frequency domain signals by applying Fourier transform.

Using these two techniques of polynomial approximation and signal decomposition with Fourier transform, our REEL algorithm transforms PDE models into decomposable PDEs with sparse value and sparse frequency domain feature functions, and then use random projection to compress the sparse signals. We then use these compressed signals to learn the PDE model parameters. We now describe our REEL learning framework in more details.

**Taylor series approximation**

For PDE models that are not decomposable into parameter functions and feature functions, we use Taylor series approximation upto certain order terms and the resulting polynomial can then be written as the decomposable form in Equation 3. Assuming that the function $F$ in Equation 1 is infinitely differentiable at $\theta = a$, and

dropping the spatial derivatives $\nabla u, \nabla^2 u, \ldots$ to avoid cumbersome notation, we can write:

$$\frac{\partial u(\vec{p}, t)}{\partial t} = F(u, \nabla u, \nabla^2 u, \ldots, \theta)$$

$$\approx F(\theta = a) + (\theta - a)\frac{\partial F}{\partial \theta}\Big|_{\theta = a}$$

$$+ \frac{1}{2!}(\theta - a)^2 \frac{\partial^2 F}{\partial \theta^2}\Big|_{\theta = a} + \ldots$$

$$= [1, (\theta - a), \frac{1}{2}(\theta - a)^2, \ldots] \begin{bmatrix} F \\ \frac{\partial F}{\partial \theta} \\ \frac{\partial^2 F}{\partial \theta^2} \\ \vdots \end{bmatrix}_{\theta = a}$$

This is the same form as in Equation 3. Approximation with Taylor expansion comes with approximation errors which we quantify with the following theorem:

**Theorem 3.1.** *(Proof in supplementary material) If $F(\theta)$ is at least $(n + 1)$ time differentiable around $\theta = a$, except possibly at $a$, then $F(\theta) = \sum_{i=0}^{n} \frac{(\theta-a)^n}{n!} \frac{\partial^n F}{\partial \theta^n}\Big|_{\theta=a} + E_n$, where the approximation error $E_n = \frac{(\theta-a)^{n+1}}{(n+1)!} \frac{\partial^n F}{\partial \theta^n}\Big|_{\theta=c}$ for a suitable $c$ in the closed range joining $\theta$ and $a$.*

Theorem 3.1 is a reformulation of Lagrange's remainder, and provides us error bound $E_n$ for n-th order Taylor series approximation. In practice, we use empirical testing to decide on the order of the polynomial.

**Signal decomposition with Fourier transform**

Algorithm 1: Value and Frequency Domain Decomposition (VFDD)

1: **Input:** Signal $s$, threshold $\beta$
2: **Output:** Frequency domain component $s_{freq}$ and value domain component $s_{val}$ of signal $s$
3: Computer $\mathcal{F}(s)$, the Discrete Fourier Transform of $s$;
4: $s_{freq} \leftarrow \mathcal{F}(s) \times \mathbf{1}_{|\mathcal{F}(s)| > \beta}$;
5: $s_{val} \leftarrow \mathcal{F}^{-1}(\mathcal{F}(s) \times \mathbf{1}_{|\mathcal{F}(s)| \leq \beta})$;

With a PDE model that is now decomposed into parameter function and feature function as in Equation 3, we now look into sparsity structure of the problem. Let $\Delta u^{GT}(t)$ be the ground truth change in system variable $u$ at time $t$. We use value and frequency domain signal decomposition as outlined in Algorithm 1 to convert dense $\Delta u^{GT}(t)$ signals into combination of sparse value and sparse frequency signals.

In Algorithm 1, given an input signal and a threshold, we first compute the discrete Fourier transform of the signal (line 3). We then separate the high coefficient frequency terms from the low ones based on the threshold (line 4). Using inverse discrete Fourier transform, we then convert the low coefficient frequency terms back to value domain (line 5). Note that both value and frequency domain components have the same size as the original signal, with a portion of them zeroed out based on the threshold, which is chosen by empirically testing few signal samples.

After value and frequency domain decomposition, $\Delta u^{GT}(t)$ is separated into sparse frequency component $\Delta u^{GT}_{freq}(t)$ and sparse value component $\Delta u^{GT}_{val}(t)$. On broad stroke, $\Delta u^{GT}_{freq}(t)$ corresponds to dense but slow background change of $u^{GT}(t)$, while $\Delta u^{GT}_{val}(t)$ corresponds to sharp but small interfacial change. In a similar way, we can also convert $W_i$, the feature function in Equation 3, into separate frequency and value domain components $W_{i(freq)}$ and $W_{i(val)}$. Then the PDE model predicted value domain system state change $\Delta u_{val}(t)$ can be computed as:

$$\Delta u_{val}(t) = dt.[\phi_1(\theta), \phi_2(\theta), \ldots, \phi_n(\theta)] \begin{bmatrix} W_{1(val)}(u) \\ W_{2(val)}(u) \\ \vdots \\ W_{n(val)}(u) \end{bmatrix}$$

The frequency domain change $\Delta u_{freq}(t)$ can be computed in similar manner. The parameter functions $\phi_i$ are same for both value domain and frequency domain components.

**Signal compression with random projection**

After signal decomposition into sparse value and sparse frequency domain components, we use one-time random projection with random matrix $\boldsymbol{P}$ to compress all the sparse signal components for both system state change (yielding $\boldsymbol{P}\Delta u^{GT}_{val}, \boldsymbol{P}\Delta u^{GT}_{freq}$) and feature functions (yielding $\boldsymbol{P}W_{i(val)}, \boldsymbol{P}W_{i(freq)}$). Then the compressed feature functions can be used to compute predicted compressed system change using Equation 4. Finally, to learn PDE model parameters $\theta$, we minimize the following loss:

$$L_{\text{REEL}}(\theta) = \sum_{i=1}^{T} ||\boldsymbol{P}\Delta u_{val}(t) - \boldsymbol{P}\Delta u^{GT}_{val}(t)||_2^2$$
$$+ \lambda ||\boldsymbol{P}\Delta u_{freq}(t) - \boldsymbol{P}u^{GT}_{freq}(t)||^2. \quad (5)$$

Here, $\lambda$ is a hyperparameter. The loss function in Equation 5 consists of two parts. The first part inside the summation penalizes the difference between the predicted compressed value domain change and ground truth compressed value domain change, while the second part penalizes the difference between the frequency domain counterparts. We can then use stochastic gradient descent to find the optimal parameters $\theta$ that minimize loss $L_{\text{REEL}}(\theta)$ in Equation 5.

**Theorem 3.2.** *(Proof in supplementary material) Suppose the projection matrix $\boldsymbol{P} = (p_{i,j})_{n \times d}$, $p_{i,j} = y_{i,j}/\sqrt{n}$. $y_{i,j}$ are sampled i.i.d. from a given distribution. $y_i^T = (y_{i,1}, \ldots, y_{i,d})$, $Y = (y_1, \ldots, y_n)^T$. $E(y_{i,j}) = 0$, $Var(y_{i,j}) = 1$. For any $x$, $\|y_i^T x\|^2 / \|x\|_2^2$ is sub-exponential with parameter $(\sigma^2, b)$. After value and frequency domain decomposition as outlined in Algorithm 1, let $\Delta u_{val}^{GT}(t)$ and $\Delta u_{val}(t)$ have at most $k_1$ non-zero elements, and all $\Delta u_{freq}^{GT}(t)$ and $\Delta u_{freq}(t)$ have at most $k_2$ non-zero elements. $2k_1 < n, 2k_2 < n$. $0 < \delta < \min\{1, \sigma^2/b\}$. We separate the loss function in value domain and in frequency domain without random projection: $L_{val} = \sum \|\Delta u_{val}(t) - \Delta u_{val}^{GT}(t)\|^2$, $L_{freq} = \sum \|\Delta u_{freq}(t) - \Delta u_{freq}^{GT}(t)\|^2$. Suppose $\theta^*$ is the optimal parameter which minimizes $L_T(\theta) = L_{value} + \lambda L_{freq}$, i.e., $\theta^* = \arg\min L_T(\theta)$. Then with probability at least $[1 - 2(12/\delta)^{2k_1} \exp(-n\delta^2/(8\sigma^2))][1 - 2(12/\delta)^{2k_2} \exp(-n\delta^2/(8\sigma^2))]$, we have:*

$$(1-\delta)^2 L_T(\theta^*) \leq L_{\text{REEL}}(\theta^*) \leq (1+\delta)^2 L_T(\theta^*). \quad (6)$$

*On the opposite side, suppose $\theta'$ is the local optimal solution found by* REEL*, with the same probability we have:*

$$(1-\delta)^2 L_T(\theta') \leq L_{\text{REEL}}(\theta') \leq (1+\delta)^2 L_T(\theta'). \quad (7)$$

In layman terms, Theorem 3.2 implies that random projection in value and frequency domain has limited effect on learning provided that the signals are sufficiently sparse after value and frequency domain decomposition, and we only require poly-logarithmic number of projected dimensions for constant factor approximation.

## 4 Related works

**Learning dynamics models.** Machine learning to learn physics dynamics models have been a popular research domain in recent years. Recently, learning Partial Differential Equations (PDEs) from data has also been studied extensively Dzeroski and Todorovski [1995], Brunton et al. [2016], Wu and Tegmark [2019], Zhang and Lin [2018], Iten et al. [2020], Cranmer et al. [2020a], Raissi et al. [2020, 2019b], Liu and Tegmark [2021], Xue et al. [2021b], Chen et al. [2018]. Many of the existing works in learning dynamics from data combines domain knowledge with machine learning models such as neural networks. Some of the notable works include (Sirignano and Spiliopoulos [2018],Raissi et al. [2019a], Lutter et al. [2018], Demeester [2019], Long et al. [2018]), Xue et al. [2021a] where neural networks have been used to solve PDEs for dynamic systems. Besides these, Han et al. [2018], Beck et al. [2019], Raissi and Karniadakis [2018], Brunton et al. [2016] are some of the other notable works in PDE solution. Neural ODEs proposed in (Chen et al. [2018]) and their variants such as Kidger et al. [2020], Lee and Parish [2021], Jia and Benson [2019], Chen et al. [2020], Yin et al. [2021] aim to learn ordinary differential equation based dynamics models.

**Learning physics models from data.** Physics model learning from data have been explored in Greydanus et al. [2019], Cranmer et al. [2020b], Lutter et al. [2018], Niu et al. [2020]. In regards to efficient methods for physics learning, Xue et al. [2021a], Sima and Xue [2021], Nasim et al. [2022],Bar-Sinai et al. [2019],Schaeffer [2017] are similar to our works in that these also aim to make the learning more computationally efficient, using locality sensitive hashing, random projection, approximate derivatives and compressed sensing. However, we introduce the frequency domain decomposition in the learning pipeline, which was not used in previous approaches to learn physics models. To make learning more efficient, previously data compression methods such as principal component analysis, low rank approximation, feature selection etc. Pruning, quantization, low rank factorization, knowledge distillation are some of the popular techniques used to compress deep neural networks previously Choudhary et al. [2020]. In our work, we focus on data compression, with the added change of representation to both value and frequency domain.

# 5 Experimental results

## Solid-state selective laser sintering

Selective laser sintering (SLS) is a widely used important manufacturing process, where laser energy is used to sinter powder particles into dense solid structures. Accurate modeling of microstructure evolution during sintering is very important for process control and optimization. For our experiment, we work with the phase field model of solid-state sintering as proposed in Zhang and Liao [2018], where a thermal model is coupled with microstructure model.

**Thermal model.** Heat diffusion from laser energy source during sintering process can be modeled with transient heat conduction equation as follows:

$$\rho \frac{\partial C_p T}{\partial T} = \nabla .(k \nabla T) + Q. \qquad (8)$$

Here, $T$ represents the temperature field in the specimen. $\rho, C_p, k$ are the density, specific heat and thermal conductivity of the material respectively and assumed to be material specific constants in our experiments. $Q$ is heat flux from the laser heat source.

**Microstructure model.** In the phase field model of sintering, the microstructure is represented with two types of field variables – conserved density field $\phi$ and non-conserved order parameters $\eta_i$ for each particle. $\phi$ takes value in the range $[0, 1]$, where 1 represent solid phase and 0 represent pores. $\eta_i$ similarly has value in range $[0, 1]$, and takes the value of 1 for a designated particle and 0 elsewhere. The driving force for sintering is the minimization of the total free energy $F$,

$$F = \int_V \left[ f(\phi, \eta_{i=1,\dots,n}) + \frac{\varepsilon_\phi}{2}|\nabla \phi|^2 + \sum_{i=1}^n \frac{\varepsilon_\eta}{2}|\nabla \eta_i|^2 \right] dV.$$

Here, $n$ is the number of particles in the system. The evolution of $\phi$ and $\eta_i$ over time are governed by Cahn-Hilliard equation and Allen-Cahn equation respectively:

$$\frac{\partial \phi}{\partial t} = \nabla \cdot (M \nabla \frac{\delta F}{\delta \phi}), \qquad \frac{\partial \eta}{\partial t} = -L \frac{\delta F}{\delta \eta}, \qquad (9)$$

where $M$ and $L$ represent atom diffusion mobility and grain boundary (GB) mobility respectively. For details of both thermal and microstructure model, we refer to the original text Zhang and Liao [2018].

**Learning objective.** Given the system states $\phi, \eta$ and $T$, we aim to learn the parameters of the selective laser sintering model, which are $\{C_p, \rho, k\}$ and parameters associated with $F, M$ and $L$ in Equation 11.

**Training and Testing.** For our experiment with sintering application, we used synthetic dataset according to the model in Zhang and Liao [2018]. We simulated microstructure evolution and heat diffusion during sintering in 2D for $N \times N$ grids, for $N = 100, 200, 300, 400, 500$ for $T = 20000$ timesteps. For training, we used data for 1000 timesteps. **As the baseline method, we used our REEL algorithm without the random projection step.** Stochastic gradient descent was used for optimization during training, and the learning rate was set by hyperparameter tuning. More details of training and testing (i.e. computing resources, code etc.) are provided in supplementary materials.

## Nanovoid evolution in materials under irradiation and high temperature

Materials under heavy irradiation and high temperature forms many types of defects. One of these defects is named *nanovoid*, which are nano-meter scale defect clusters, formed by the accumulation of vacancy defect in crystal lattice. Such void defects greatly affects material degradation over time. Modeling the evolution dynamics of such defects are very important in designing sustainable materials that can withstand extreme environments such as inside a nuclear reactor.

In the phase field model for nanovoid defect evolution in engineering materials, the state of a system is described by 3 phase field variables – $c_v, c_i$ and $\eta$ at each point in space and time. $c_v$ and $c_i$ represents the percentage of vacancy defects and interstitial defects respectively, in the crystal lattice, while $\eta$ is an order parameter distinguishing between different phases of the material. For our experiment, we use the phase field PDE model proposed in Millett et al. [2011] which describe the time evolution of phase field variables with with Cahn-Hilliard equation and Allen-Cahn equation as shown in Equation 11. For details of the phase field model, we refer to original text Millett et al. [2011].

**Training and Testing.** We used a synthetic dataset generated according to the phase field model in Millett et al. [2011] for our nanovoid experiments. For baseline, we used RAPID-PDE Nasim et al. [2022] with two different compression levels – 1) no compression and 2) compression to $10\%$ of original data dimension. **We used RAPID-PDE as baseline for nanovoid only, since the nanovoid PDEs are decomposable, while the sintering model PDEs are not.**

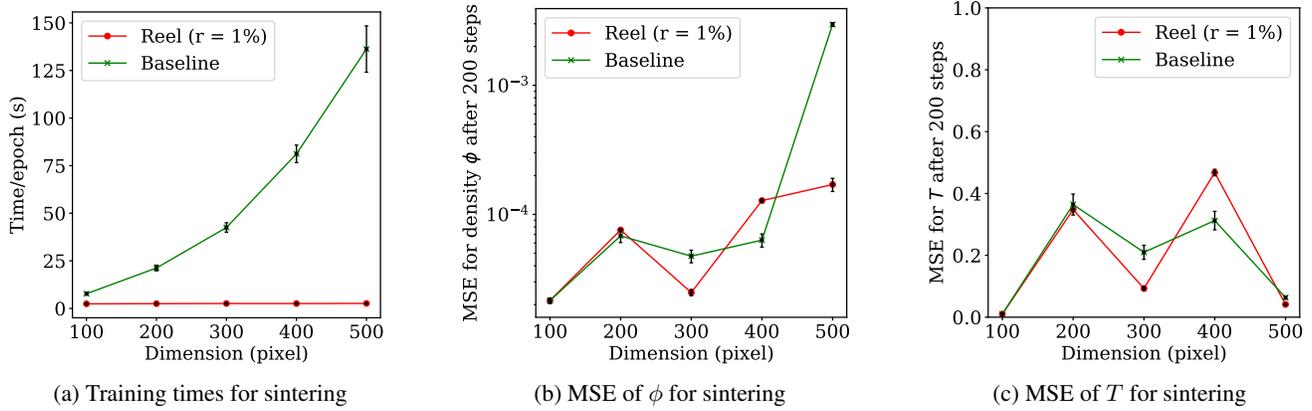

Figure 2: Our REEL algorithm saves 70% to 98% of training times for learning PDE physics models, while preserving very high accuracy of the learned models comparable to baseline (we used REEL algorithm **without** compression as our baseline for sintering). (a) Training times for sintering PDE model. Here, $r$ denotes the compression level used in REEL, $r = 1\%$ means the data was compressed to $1\%$ of the original dimension. (b-c) Mean squared error (MSE) for density ($\phi$) and temperature ($T$) field variable in sintering model. MSE for a model was computed by performing simulation with the model for 200 timesteps, and then comparing the simulation output and ground truth. MSE of $\phi$ and $T$ are very small and comparable between our REEL and baseline, and the simulation results are practically indistinguishable as shown in Figure 3.

## Results

**Faster learning of PDE physics models.** Our REEL algorithm leads to faster training of PDE physics models compared to the baseline learning methods. As shown in Figure 2a, our REEL algorithm can reduce training time for sintering PDE model by $70-98\%$. For nanovoid PDE model training, we see a similar level of computational saving ($\approx 70\%$ reduction of training time) when no compression is used. Added to the benefit of reduced data dimension from compression, the data size also becomes small enough to perform the training epochs in memory, while for baseline method with no compression, the training epochs has to be performed with data stored in disks. We tried to train the baseline model by keeping the entire training data in memory, however the training data was too large to fit in memory. More details on experiment setups and infrastructure are provided in supplementary material.
**Highly accurate learned physics models.** Our REEL training algorithm leads to highly accurate PDE physics models. We used simulation to qualitatively evaluate the models trained with our algorithm, and the simulation results are presented in Figure 3. Overall, we find that the models trained with our REEL algorithm produce very similar dynamics as ground truth. For sintering, the baseline method with no compression also learn very accurate dynamics, however requires very high model training time as shown in Figure 2a. For quantitative evaluation, we simulated all our trained models for 200 steps using the same initial condition. This was repeated 200 times with different initial conditions, which were unseen during training. Afterwards, we computed the mean squared error (MSE) between simulation results and ground truth as shown in Figure 2b and Figure 2c. We can see that even with data dimension reduced to $1\%$ of the original size, our REEL trained models show comparable very small error similar to baseline models. For nanovoid model, we used RAPID-PDE with different compression levels as our baseline. With data compressed to $10\%$ of original size, REEL algorithm can capture the shrinking dynamics of void evolution as shown in Figure 3b (4-th row), while baseline RAPID-PDE algorithm results in the disintegration of void as seen in Figure 3b (3-rd row). More details of evaluation result with nanovoid

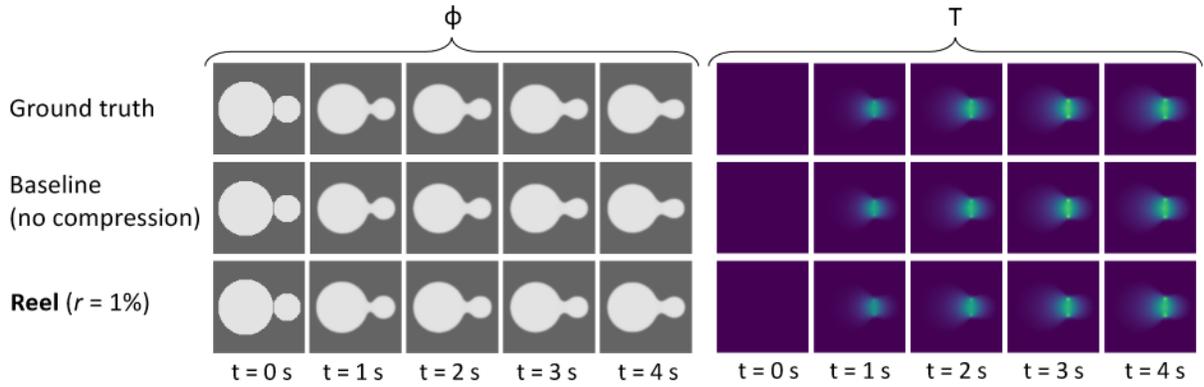

(a) Simulation of particle density ($\phi$) and temperature ($T$) change during laser sintering of powder particles

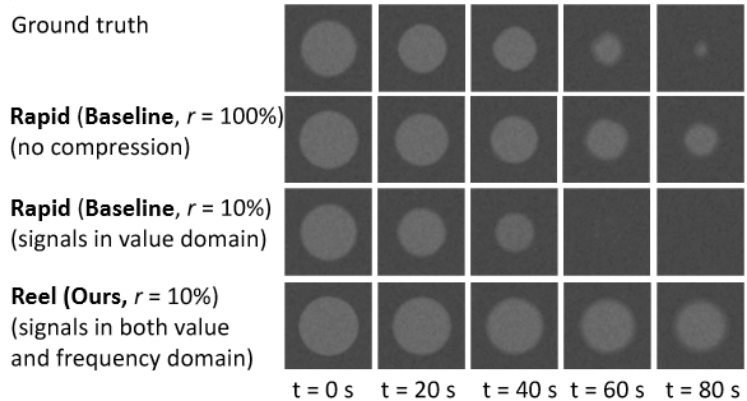

(b) Simulation of nanovoid evolution under extreme conditions

Figure 3: REEL can learn very accurate physics models governing sintering of powder particles and nanovoid defect evolution. (a) Simulation of mircrostructure evolution and heat diffusion during solid-state selective laser sintering (b) Simulation of nanovoid defect evolution in materials under extreme heat and irradiation. Here, $r$ represents the level of data compression. For example, $r = 1\%$ means data has been compressed to $1\%$ of original size. Our REEL algorithm can capture the slow shrinking dynamics of nanovoids as seen in the simulation (4-th row in Figure 3b), while with baseline RAPID-PDE, the void disintegrates (3-rd row in Figure 3b). Our REEL method is applicable to both PDE models, while RAPID-PDE algorithm, which is used as baseline for nanovoid experiments, cannot be applied to PDEs of sintering phase field model.

PDE model is provided in supplementary material.

# 6 Conclusion

In this paper, we present REEL, an efficient algorithm to learn partial differential equations from experimental data. Our acceleration is based on decomposing the spatial and temporal updates into sparse signals in the value and frequency domains. REEL also uses Taylor series expansion to approximate PDE models in a decomposable form. Random projection is then applied to compress the sparse signals in the value and frequency domains as a preprocessing step. Learning is carried out entirely in the compressed space. Our method is applicable to a wide range of PDE models where the spatial and temporal updates are made of slow and wide-range changes in addition to rapid changes in the interfacial regions. Empirically, we show that our algorithm can lead to faster learning of physics models, and the learned models exhibit reasonably high accuracy on testing.

## Acknowledgements

This research was supported by NSF grants CCF-1918327.

# 7 Supplementary Material

## Code

The code for experiments will be released for public use in future.

## Theorems and proofs

**Theorem 3.1.** *If $F(\theta)$ is at least $(n+1)$ time differentiable around $\theta = a$, except possibly at $a$, then $F(\theta) = \sum_{i=0}^{n} \frac{(\theta-a)^n}{n!} \frac{\partial^n F}{\partial \theta^n}\Big|_{\theta=a} + E_n$, where the approximation error $E_n = \frac{(\theta-a)^{n+1}}{(n+1)!} \frac{\partial^n F}{\partial \theta^n}\Big|_{\theta=c}$ for a suitable $c$ in the closed range joining $\theta$ and $a$.*

*Proof.* Theorem 3.1 is easily obtained from the Taylor series formula with Lagrange's remainder, and we refer to existing text for the proof (Section 7.7 Apostol [1991]). □

**Theorem 3.2.** *Suppose the projection matrix $\boldsymbol{P} = (p_{i,j})_{n \times d}$, $p_{i,j} = y_{i,j}/\sqrt{n}$. $y_{i,j}$ are sampled i.i.d. from a given distribution. $y_i^T = (y_{i,1}, \ldots, y_{i,d})$, $Y = (y_1, \ldots, y_n)^T$. $E(y_{i,j}) = 0$, $Var(y_{i,j}) = 1$. For any $x$, $||y_i^T x||^2/||x||_2^2$ is sub-exponential with parameter $(\sigma^2, b)$. After value and frequency domain decomposition as outlined in Algorithm 1, let $\Delta u_{val}^{GT}(t)$ and $\Delta u_{val}(t)$ have at most $k_1$ non-zero elements, and all $\Delta u_{freq}^{GT}(t)$ and $\Delta u_{freq}(t)$ have at most $k_2$ non-zero elements. $2k_1 < n, 2k_2 < n$. $0 < \delta < \min\{1, \sigma^2/b\}$. We separate the loss function in value domain and in frequency domain without random projection: $L_{val} = \sum ||\Delta u_{val}(t) - \Delta u_{val}^{GT}(t)||^2$, $L_{freq} = \sum ||\Delta u_{freq}(t) - \Delta u_{freq}^{GT}(t)||^2$. Suppose $\theta^*$ is the optimal parameter which minimizes $L_T(\theta) = L_{val} + \lambda L_{freq}$, i.e., $\theta^* = \arg\min L_T(\theta)$. Then with probability at least $[1 - 2(12/\delta)^{2k_1} \exp(-n\delta^2/(8\sigma^2))][1 - 2(12/\delta)^{2k_2} \exp(-n\delta^2/(8\sigma^2))]$, we have:*

$$(1-\delta)^2 L_T(\theta^*) \leq L_{\text{REEL}}(\theta^*) \leq (1+\delta)^2 L_T(\theta^*). \quad (8)$$

*On the opposite side, suppose $\theta'$ is the local optimal solution found by* REEL*, with the same probability we have:*

$$(1-\delta)^2 L_T(\theta') \leq L_{\text{REEL}}(\theta') \leq (1+\delta)^2 L_T(\theta'). \quad (9)$$

*Proof.* Before the proof, we provide definition of sub-exponential, and an auxiliary lemma that we will use in our proof.

**Definition.** $X$ is a random variable, $E(x) = \mu$. $M_{x-\mu}(\lambda) = E[exp(\lambda(X-\mu))]$ is the moment generating function of $X - \mu$. $X$ is sub-exponential with parameter $(\sigma^2, b)$ if for all $|\lambda| < 1/b$, $\ln M_{x-\mu}(\lambda) \leq \lambda^2 \sigma^2 / 2$.

We will use the following auxiliary lemma from Nasim et al. [2022] in our proof.

**Lemma 3.3.** *(Source: Theorem 5.1 Nasim et al. [2022]) Suppose the projection matrix $P = (p_{i,j})_{n \times d}$, $p_{i,j} = y_{i,j}/\sqrt{n}$. $y_{i,j}$ are sampled i.i.d. from a given distribution. $y_i^T = (y_{i,1}, \ldots, y_{i,d})$, $Y = (y_1, \ldots, y_n)^T$. $E(y_{i,j}) = 0$, $Var(y_{i,j}) = 1$. For any $x$, $||y_i^T x||^2 / ||x||_2^2$ is sub-exponential with parameter $(\sigma^2, b)$. All $\Delta u'_{t_i}$ and $\Delta u_{t_i}$ have at most $k$ non-zero elements, $2k < n$. $0 < \delta < \min\{1, \sigma^2/b\}$. Suppose $\theta^*$ is the optimal parameter which minimizes $L(\theta)$, i.e., $\theta^* = \arg\min L(\theta)$. Then with probability at least $1 - 2(12/\delta)^{2k} \exp(-n\delta^2/(8\sigma^2))$, we have:*

$$(1-\delta)^2 L(\theta^*) \leq L'(\theta^*) \leq (1+\delta)^2 L(\theta^*).$$

*On the opposite side, suppose $\theta'$ is the local optimal solution found by* RAPID-PDE*, with the same probability we have:*

$$(1-\delta)^2 L(\theta') \leq L'(\theta') \leq (1+\delta)^2 L(\theta').$$

Now we begin our proof for Theorem 3.2. For notation convenience, lets split the two part loss function $L_{\text{REEL}}$ as follows:

$$L_{\text{REEL}(val)}(\theta) = \sum ||\boldsymbol{P}\Delta u_{val}(t) - \boldsymbol{P}\Delta u_{val}^{GT}(t)||_2^2$$
$$L_{\text{REEL}(freq)}(\theta) = \sum ||\boldsymbol{P}\Delta u_{freq}(t) - \boldsymbol{P}u_{freq}^{GT}(t)||^2.$$

Using Lemma 3.3, we can see that the following holds for $L_{\text{REEL}(val)}$ with probability at least $1 - 2(12/\delta)^{2k_1} \exp(-n\delta^2/(8\sigma^2))$,

$$(1-\delta)^2 L_{val}(\theta^*) \leq L_{\text{REEL}(val)}(\theta^*) \leq (1+\delta)^2 L_{val}(\theta^*).$$

Similarly, for $L_{freq}$, the following holds with probability at least $1 - 2(12/\delta)^{2k_2} \exp(-n\delta^2/(8\sigma^2))$,

$$(1-\delta)^2 L_{freq}(\theta^*) \leq L_{\text{REEL}(freq)}(\theta^*) \leq (1+\delta)^2 L_{freq}(\theta^*).$$

Therefore, with probability at least $[1 - 2(12/\delta)^{2k_1} \exp(-n\delta^2/(8\sigma^2))][1 - 2(12/\delta)^{2k_2} \exp(-n\delta^2/(8\sigma^2))]$, the following holds:

$$(1-\delta)^2 [L_{val}(\theta^*) + \lambda L_{freq}(\theta^*)]$$
$$\leq [L_{\text{REEL}(val)}(\theta^*) + \lambda L_{\text{REEL}(freq)}(\theta^*)]$$
$$\leq (1+\delta)^2 [L_{val}(\theta^*) + \lambda L_{freq}(\theta^*)]$$
$$\Rightarrow (1-\delta)^2 L_T(\theta^*) \leq L_{\text{REEL}}(\theta^*) \leq (1+\delta)^2 L_T(\theta^*).$$

In a similar manner, we can prove the second part of our Theorem 3.2, that if $\theta'$ is the local optimal solution found by REEL, then with the same probability we have:

$$(1-\delta)^2 L_T(\theta') \leq L_{\text{REEL}}(\theta') \leq (1+\delta)^2 L_T(\theta').$$

□

## Solid-state selective laser sintering

For our experiment, we work with the phase field model of solid-state sintering as proposed in Zhang and Liao [2018], where a thermal model is coupled with microstructure model.

**Thermal model.** Heat diffusion from laser energy source during sintering process can be modeled with transient heat conduction equation as follows:

$$\rho \frac{\partial C_p T}{\partial T} = \nabla \cdot (k \nabla T) + Q. \quad (10)$$

Here, $T$ represents the temperature field in the specimen. $\rho, C_p, k$ are the density, specific heat and thermal conductivity of the material respectively, and Q is heat flux from the laser heat source. We model the heat flux to have a Gaussian profile as follows:

$$Q = \frac{2\Gamma}{\pi \omega^2} e^{-\frac{x^2+y^2}{2\omega^2}}$$

Here, $\Gamma$ denotes the laser power and $\omega$ denotes effective laser spot radius. For simplicity, we assume the laser is stationary hovering over the specimen.

**Microstructure model.** In the phase field model of sintering, the microstructure is represented with two types of field variables – conserved density field $\phi$ and non-conserved order parameters $\eta_i$ for each particle. $\phi$ takes value in the range $[0, 1]$, where 1 represent solid phase and 0 represent pores. $\eta_i$ similarly has value in range $[0, 1]$, and takes the value of 1 for a designated particle and 0 elsewhere. The driving force for sintering is the minimization of the total free energy $F$,

$$F = \int_V \left[ f(\phi, \eta_{i=1,\ldots,n}) + \frac{\varepsilon_\phi}{2}|\nabla\phi|^2 + \sum_{i=1}^n \frac{\varepsilon_\eta}{2}|\nabla\eta_i|^2 \right] \mathrm{d}V.$$

Here, $n$ is the number of particles in the system. The evolution of $\phi$ and $\eta_i$ over time are governed by Cahn-Hilliard equation and Allen-Cahn equation respectively:

$$\frac{\partial \phi}{\partial t} = \nabla \cdot (M \nabla \frac{\delta F}{\delta \phi}), \qquad \frac{\partial \eta}{\partial t} = -L \frac{\delta F}{\delta \eta}, \quad (11)$$

where $M$ and $L$ represent atom diffusion mobility and grain boundary (GB) mobility respectively.

Mass transport during sintering involves multiple diffusion mechanisms such as bulk diffusion, surface diffusion, vapor transport, GB diffusion etc., and we can model these different mechanisms by writing M as follows:

$$M = M_{vol} h(\phi) + M_{vap}[1 - h(\phi)] + M_{surf}\phi[1-\phi] + M_{GB} \sum_{i \neq j} \eta_i \eta_j$$

where $h(\phi) = \phi^3(15 - 10\phi + 6\phi^2)$, and $M_{vol}, M_{vap}, M_{surf}, M_{GB}$ are mobility coefficients for volume, vapor, surface and grain boundary diffusion. The mobility coefficients for different diffusion modes are related to the diffusion coefficients as follows:

$$M_l = \frac{D_l V_m}{k_B T}$$

where $l \in \{vol, vap, surf, GB\}$, $D_l$ is the diffusion coefficient for particular diffusion mode, $V_m$ is the molar volume, $k_B$ is the Boltzman constant and $T$ is the temperature field as obtained from the thermal model. The value of diffusion coefficients are estimated from the Arrhenius equation as:

$$D_l = D_{l0} e^{-\frac{Q_l}{k_B T}}$$

Here, $D_{l0}$ and $Q_l$ denote prefactor and activation energy. In our experiments, we learn the parameters $V_m, \varepsilon_\phi, \varepsilon_\eta$, and $D_{l0}, Q_l$ for $l \in \{vol, vap, surf, GB\}$.

**Taylor series approximation for sintering model.** The transient heat diffusion relation (Equation 10) and Allen-Cahn equation for non-conserved phase fields (Equation 11) are readily decomposable into inner products of feature functions. Here, we show how Cahn-Hilliard equation for conserved phase fields (Equation 11 can be approximated with Taylor series and transformed into decomposable format. From Equation 11:

$$\begin{aligned}
\frac{\partial \phi}{\partial t} &= \nabla \cdot (M \nabla \frac{\delta F}{\delta \phi}) \\
&= (M_{vol} h(\phi) + M_{vap}[1 - h(\phi)] + M_{surf}\phi[1-\phi] \\
&\quad + M_{GB} \sum_{i \neq j} \eta_i \eta_j) \nabla^2 \frac{\delta F}{\delta \phi} \\
&= \frac{D_{vol0} e^{-\frac{Q_{vol}}{k_B T}} V_m}{k_B T} h(\phi) \nabla^2 \frac{\delta F}{\delta \phi} \\
&\quad + \frac{D_{vap0} e^{-\frac{Q_{vap}}{k_B T}} V_m}{k_B T}[1 - h(\phi)] \nabla^2 \frac{\delta F}{\delta \phi} \\
&\quad + \frac{D_{surf0} e^{-\frac{Q_{surf}}{k_B T}} V_m}{k_B T} \phi(1-\phi) \nabla^2 \frac{\delta F}{\delta \phi} \\
&\quad + \frac{D_{GB0} e^{-\frac{Q_{GB}}{k_B T}} V_m}{k_B T} \sum_{i \neq j} \eta_i \eta_j \nabla^2 \frac{\delta F}{\delta \phi} \quad (12)
\end{aligned}$$

Equation 12 consists of 4 terms added together, with the activation energy parameters $Q_l$ in the exponential power. For the first term, using Taylor series, we can write:

$$\begin{aligned}
&\frac{D_{vol0} e^{-\frac{Q_{vol}}{k_B T}} V_m}{k_B T} h(\phi) \nabla^2 \frac{\delta F}{\delta \phi} \\
&= D_{vol0}[1 - \frac{Q_{vol}}{k_B T} + \frac{1}{2!}(\frac{Q_{vol}}{k_B T})^2 \\
&\quad - \frac{1}{3!}(\frac{Q_{vol}}{k_B T})^3 + \ldots] h(\phi) \nabla^2 \frac{\delta F}{\delta \phi} \\
&= [D_{vol0}, D_{vol0} Q_{vol}, \ldots] \begin{bmatrix} h(\phi) \nabla^2 \frac{\delta F}{\delta \phi} \\ -\frac{h(\phi)}{k_B T} \nabla^2 \frac{\delta F}{\delta \phi} \\ \vdots \end{bmatrix} \quad (13)
\end{aligned}$$

Using upto $n^{th}$ term in the Taylor series, we can thus decompose all the terms in Equation 12, and hence Equation 11 into inner products of parameter functions and feature functions. The second equation $\frac{\partial \eta}{\partial t} = -L \frac{\delta F}{\delta \eta}$ is already decomposable and hence we do not use Taylor series approximation.

# Nanovoid evolution in materials under irradiation and high temperature

Materials under heavy irradiation and high temperature forms many types of defects. One of these defects is named *nanovoid*, which correspond to missing atoms in crystal lattice. Such void defects greatly affects material degradation over time, and modeling the evolution dynamics of such defects are very important in designing sustainable materials. For our experiment, we use the phase field model of nanovoid defect evolution proposed in Millett et al. [2011].

In the phase field model for nanovoid defect evolution in engineering materials, the state of a system is described by 3 phase field variables – $c_v, c_i$ and $\eta$ at each point in space and time. $c_v$ and $c_i$ represents the percentage of vacancy defects and interstitial defects respectively, in the crystal lattice, while $\eta$ is an order parameter distinguishing between different phases of the material. The evolution of the 3 phase-field variables minimizes the total free energy $F$ of the system, where

$$F = N \int_V \left[ h(\eta) f^s(c_v, c_i) + j(\eta) f^v(c_v, c_i) + \frac{\kappa_v}{2}|\nabla c_v| + \frac{\kappa_i}{2}|\nabla c_i| + \frac{\kappa_\eta}{2}|\nabla \eta| \right] dV$$

where $h(\eta) = (\eta - 1)^2, j(\eta) = \eta^2$. The time evolution of phase field variables with modified Cahn-Hilliard equation and Allen-Cahn equation as follows:

$$\frac{\partial c_v}{\partial t} = \nabla \cdot (M_v \nabla \frac{1}{N} \frac{\delta F}{\delta c_v}) + \xi(\mathbf{r}, t) + P_v(\mathbf{r}, t) - R_{iv}(\mathbf{r}, t),$$
$$\frac{\partial c_i}{\partial t} = \nabla \cdot (M_i \nabla \frac{1}{N} \frac{\delta F}{\delta c_i}) + \zeta(\mathbf{r}, t) + P_i(\mathbf{r}, t) - R_{iv}(\mathbf{r}, t),$$
$$\frac{\partial \eta}{\partial t} = -L \frac{\delta F}{\delta \eta} + P_{v,i}(\mathbf{r}, t).$$

Here, $\mathbf{r}$ represents space coordinates. $M_v(M_i)$ and $\frac{1}{N}\frac{\delta F}{\delta c_v}(\frac{1}{N}\frac{\delta F}{\delta c_i})$ represents mobility and chemical potentials of vacancies (interstitials) respectively. $\xi$ and $\zeta$ represents thermal fluctuations of vacancy and interstitial concentration fields. $P_v, P_i$ and $P_{v,i}$ represent production of vacancies and interstitials due to irradiation. $R_{iv}$ represents mutual annihilation of vacancies-interstitials. In our experiment, we assume that we have access to the ground truth thermal fluctuations and the production terms for vacancies and interstitials. We then aim to learn the scalar parameters $L, \kappa_v, \kappa_i, \kappa_\eta$ and the scalar parameters associated with $P_i, P_v, \xi, \zeta, R_{iv}$ and $F$.

# Signal decomposition with Fourier transform

**Sintering model.** The update signals in the Sintering PDE model are the temporal updates of field variables density $\phi$, order parameter $\eta$ and temperature $T$. We see that the updates of $\phi$ and $\eta$ are sparse in value domain, while the updates of $T$ is dense in value domain, and sparse in frequency domain. Therefore, during signal decomposition, we keep $\phi, \eta$ changes and all the associated feature functions in value domain, while for $T$, we keep the changes and features in frequency domain.

**Nanovoid evolution model.** The upate signals in the nanovoid evolution model are the temporal updates of $c_v, c_i$ and $\eta$. We see that all these phase fields have sharp change at the void defect boundary, with small magnitude changes in the background. Therefore, we split all the update signals $c_v, c_i, \eta$ and features into sparse value and sparse frequency components using our value and frequency decomposition algorithm. We then use random projection to compress all the sparse signals and learn the phase field model parameter in the compressed space.

# Computing Infrastructure

We used a shared server with two 64-core AMD Epyc 7662 "Rome" processors and 256 GB memory per server node, without any GPUs. For software tools, we used CentOS 7 Operating System, Python 3.8.10, Pytorch 1.10.0. for our computation.

# Notes on implementation

All the experimental results provided in the paper are aggregates of multiple monte carlo simulations. Each model was trained with different random initialization, and we provide the mean and standard deviations as errorbars in our plots. For Taylor approximation, we used upto 4-th order derivative approximation. For each model training, the learning rate was set by grid search out of $\{10^{-1}, 10^{-2}, 10^{-3}, 10^{-4}, 10^{-5}\}$. The elements of random projection matrix $P$ was sampled from standard normal distribution with 0 mean and unit variance.